\documentclass[11pt,letterpaper]{article}
\usepackage{emnlp2017}
\usepackage{times}
\usepackage{latexsym}
\usepackage{amsmath,amsfonts}
\usepackage{array}
\usepackage{breqn}
\usepackage{enumerate}
\usepackage{amsthm}
\usepackage{tikz}
\usepackage{diagbox}

\usepackage{hyperref}

\newcolumntype{C}[1]{>{\centering\let\newline\\\arraybackslash\hspace{0pt}}m{#1}}

\renewcommand{\Re}{{\mathbb R}}
\newcommand{\mat}[1]{{\mathbf #1}}
\newcommand{\x}{{\mathbf x}}
\newcommand{\z}{{\mathbf z}}

\newcommand{\m}{{\mathbf m}}
\newcommand{\cc}{{\mathbf c}}

\newcommand{\e}{{\textbf{e}}}

\DeclareMathOperator*{\argmin}{arg\,min}

\emnlpfinalcopy

\def\emnlppaperid{***}

\title{Deep Joint Entity Disambiguation with Local Neural Attention}

\author{Octavian-Eugen Ganea \and Thomas Hofmann\\
  Department of Computer Science \\ ETH Zurich \\
  {\tt \{octavian.ganea,thomas.hofmann\}@inf.ethz.ch}}

\date{}

\begin{document}

\maketitle

\begin{abstract}
We propose a novel deep learning model for joint document-level entity disambiguation, which leverages learned neural representations. Key components are entity embeddings, a neural attention mechanism over local context windows, and a differentiable joint inference stage for disambiguation. Our approach thereby combines benefits of deep learning with more traditional approaches such as graphical models and probabilistic mention-entity maps. Extensive experiments show that we are able to obtain competitive or state-of-the-art accuracy at moderate computational costs.
\end{abstract}

\section{Introduction}
Entity disambiguation (ED) is an important stage in text understanding which automatically resolves references to entities in a given knowledge base (KB). This task is challenging due to the inherent ambiguity between surface form mentions such as names and the entities they refer to. This many-to-many ambiguity can often be captured partially by name-entity co-occurrence counts extracted from entity-linked corpora.  

ED research has largely focused on two types of contextual information for disambiguation: \textit{local} information based on words that occur in a context window around an entity mention, and, \textit{global} information, exploiting document-level coherence of the referenced entities. Many state-of-the-art methods aim to combine the benefits of both, which is also the philosophy we follow in this paper. What is specific to our approach is that we use embeddings of entities as a common representation to assess local as well as global evidence. 

In recent years, many text and language understanding tasks have been advanced by neural network architectures. However, despite recent work, competitive ED systems still largely employ manually designed features. Such features often rely on domain knowledge and may fail to capture all relevant statistical dependencies and interactions. The explicit goal of our work is to use deep learning in order to learn basic features and their combinations from scratch. To the best of our knowledge, our approach is the first to carry out this program with full rigor.

\section{Contributions and Related Work}

There is a vast prior research on entity disambiguation, highlighted by~\cite{ji2016}. We will focus here on a discussion of our main contributions in relation to prior work. 

\noindent{\bf Entity Embeddings.}
We have developed a simple, yet effective method to embed entities and words in a common vector space. This follows the popular line of work on word embeddings, e.g.~\cite{mikolov2013distributed,pennington2014glove}, which was recently extended to entities and ED by~\cite{yamada2016joint,fang2016entity,zwicklbauer2016robust,huang2015leveraging}. In contrast to the above methods that require data about entity-entity co-occurrences which often suffers from sparsity, we rather bootstrap entity embeddings from their canonical entity pages and local context of their hyperlink annotations. This allows for more efficient training and alleviates the need to compile co-linking statistics. These vector representations are a key component to avoid hand-engineered features, multiple disambiguation steps, or the need for additional \textit{ad hoc} heuristics when solving the ED task.

\noindent{\bf Context Attention.}
We present a novel attention mechanism for local ED. Inspired by memory networks of~\cite{sukhbaatar2015end} and insights of~\cite{lazic2015plato}, our model deploys attention to select words that are informative for the disambiguation decision. A learned combination of the resulting context-based entity scores and a mention--entity prior yields the final local scores.  Our local model achieves better accuracy than the local probabilistic model of~\cite{ganea2016probabilistic}, as well as the feature-engineered local model of~\cite{globerson2016collective}.  As an added benefit, our model has a smaller memory footprint and it's very fast for both training and testing. 

There have been other deep learning approaches to define local context models for ED. For instance~\cite{francis2016capturing,he2013learning} use convolutional neural networks (CNNs) and stacked denoising auto-encoders, respectively, to learn representations of textual documents and canonical entity pages. Entities for each mention are locally scored based on cosine similarity with the respective document embedding. In a similar local setting,~\cite{sun2015modeling} embed mentions, their immediate contexts and their candidate entities using word embeddings and CNNs. However, their entity representations are restrictively built from entity titles and entity categories only. Unfortunately, the above models are rather 'black-box' (as opposed to ours which reveals the attention focus) and were never extended to perform joint document disambiguation. 

\noindent{\bf Collective Disambiguation.} Last, a novel deep learning architecture for global ED is proposed. Mentions in a document are resolved jointly, using a conditional random field~\cite{lafferty2001conditional} with parametrized potentials. We suggest to learn the latter by casting loopy belief propagation (LBP)~\cite{murphy1999loopy} as a rolled-out deep network. This is inspired by similar approaches in computer vision, e.g.~\cite{domke2013learning}, and allows us to backpropagate through the (truncated) message passing, thereby optimizing the CRF potentials to work well in conjunction with the inference scheme. Our model is thus trained end-to-end with the exception of the pre-trained word and entity embeddings. 
Previous work has investigated different approximation techniques, including: random graph walks~\cite{guorobust}, personalized PageRank~\cite{pershina2015personalized}, inter-mention voting~\cite{ferragina2010tagme}, graph pruning~\cite{hoffart2011robust}, integer linear programming~\cite{cheng2013relational}, or ranking SVMs~\cite{ratinov2011local}. Mostly connected to our approach is~\cite{ganea2016probabilistic} where  LBP is used for inference (but not learning) in a probabilistic graphical model and~\cite{globerson2016collective} where a single round of message passing with attention is performed. To our knowledge, we are one of the first to investigate  differentiable message passing for NLP problems.


\section{Learning Entity Embeddings}\label{ent_vecs}
In a first step, we propose to train entity vectors that can be used for the ED task (and potentially for other tasks). These embeddings compress the semantic meaning of entities and drastically reduce the need for manually designed features or co-occurrence statistics.

Entity embeddings are bootstrapped from word embeddings and are trained independently for each entity. A few arguments motivate this decision: (i) there is no need for entity co-occurrence statistics that suffer from sparsity issues and/or large memory footprints; (ii) vectors of entities in a subset domain of interest can be trained separately, obtaining potentially significant speed-ups and memory savings that would otherwise be prohibitive for large entity KBs;\footnote{Notably useful with (limited memory) GPU hardware.} (iii) entities can be easily added in an incremental manner, which is important in practice; (iv) the approach extends well into the tail of rare entities with few linked occurrences; (v) empirically, we achieve better quality compared to methods that use entity co-occurrence statistics.

Our model embeds words and entities in the same low-dimensional vector space in order to exploit geometric similarity between them. We start with a pre-trained word embedding map $\x : \mathcal{W} \rightarrow \Re^d$ that is known to encode semantic meaning of words $w \in \mathcal W$; specifically we use word2vec pre-trained vectors~\cite{mikolov2013distributed}. We extend this map to entities $\mathcal E$, i.e.~$\x : \mathcal{E} \rightarrow \Re^d$, as  described below.

We assume a generative model in which words that co-occur with an entity $e$ are sampled from a conditional distribution $p(w | e)$ when they are generated. Empirically, we collect word-entity co-occurrence counts $\#(w, e)$ from two sources: (i) the canonical KB description page of the entity (e.g. entity's Wikipedia page in our case), and (ii) the windows of fixed size surrounding mentions of the entity in an annotated corpus (e.g. Wikipedia hyperlinks in our case). These counts define a practical approximation of the above word-entity conditional distribution, i.e. ${\hat{p}(w | e) \propto \#(w,e)}$. We call this the "positive" distribution of words related to the entity. Next, let $q(w)$ be a generic word probability distribution which we use for sampling "negative" words unrelated to a specific entity. As in~\cite{mikolov2013distributed}, we choose a smoothed unigram distribution $q(w) = \hat{p}(w)^{\alpha}$ for some $\alpha \in (0,1)$. The desired outcome is that vectors of positive words are closer (in terms of dot product) to the embedding of entity $e$ compared to vectors of random words. Let $w^+ \!\sim \hat p(w|e)$ and  $w^- \sim q(w)$. Then, we use a max-margin objective to infer the optimal embedding for entity $e$: 
\begin{align}
& J(\z; e) := \mathbb{E}_{w^+|e} \; \mathbb{E}_{w^-} \left[ h\left( \z;w^+,w^- \right) \right]  \nonumber \\
& h(\z;w,v) := \left[\gamma - \langle \z, \x_{w}-  \x_{v} \rangle \right]_+ \\
& \x_e :=  \argmin_{\z: \|\z\|=1} J(\z; e) 
\nonumber
\end{align}
where  $\gamma>0$ is a margin parameter and $[\cdot]_+$ is the ReLU function. The  above loss is optimized using stochastic gradient descent with projection  over sampled pairs $(w^+,w^-)$. Note that the entity vector is directly optimized on the unit sphere which is important in order to obtain qualitative embeddings.

We empirically assess the quality of our entity embeddings on entity similarity and ED tasks as detailed in Section~\ref{sec:exp} and Appendix A. The technique described in this section can also be applied, in principle, for computing embeddings of general text documents, but a comparison with such methods is left as future work.


\section{Local Model with Neural Attention}\label{local}

\begin{figure*}[t]
\hspace{3cm} \includegraphics[width=0.72\textwidth]{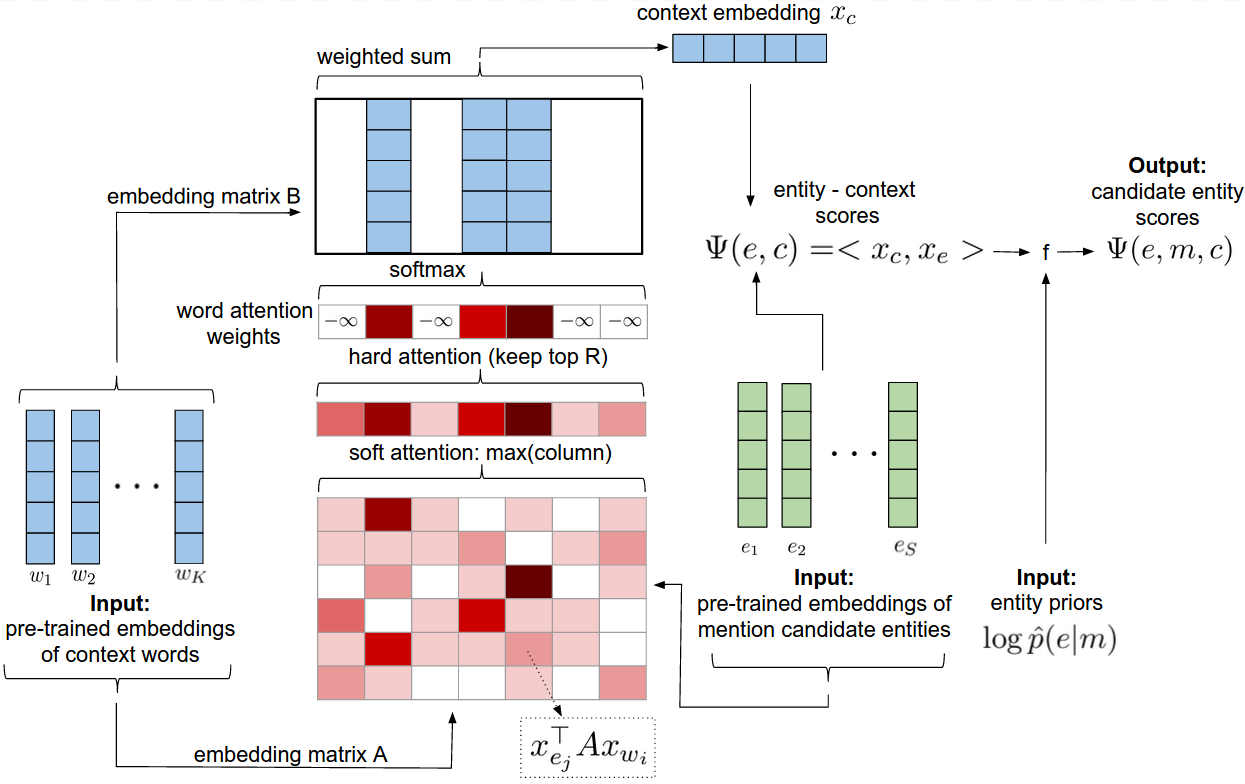}
\caption{Local model with neural attention. Inputs: context word vectors, candidate entity priors and embeddings. Outputs: entity scores. All parts are differentiable and trainable with \mbox{backpropagation}.}
\label{fig:local-model}
\end{figure*}

We now explain our local ED approach that uses word and entity embeddings to steer a neural attention mechanism. We build on the insight that only a few context words are informative for resolving an ambiguous mention, something that has been exploited before in~\cite{lazic2015plato}. Focusing only on those words helps reducing noise and improves disambiguation.~\cite{yamada2016joint} observe the same problem and adopt the restrictive strategy of removing all non-nouns. Here, we assume that a context word may be relevant, if it is strongly related to at least one of the entity candidates of a given mention.

\noindent{\bf Context Scores.}

Let us assume that we have computed a mention--entity prior $\hat p(e|m)$ (procedure detailed in Section~\ref{candidate}). In addition, for each mention $m$, a pruned candidate set $\Gamma(m)$ of at most S entities has been identified. Our model, depicted in Figure~\ref{fig:local-model}, computes a score for each $e \in \Gamma(m)$ based on the K-word local context $c= \{w_1,\ldots,w_K\}$ surrounding $m$, as well as on the prior. It is a composition of differentiable functions, thus it is smooth from input to output, allowing us to easily compute gradients and backpropagate through it. 

Each word $w \in c$ and entity $e \in \Gamma(m)$ is mapped to its embedding via the pre-trained map $\x$ (cf.~Section~\ref{ent_vecs}). We then compute an unnormalized support score for each word in the context as follows:
\begin{align}
u(w) = \max_{e \in \Gamma(m)} \x_{e}^{\top} \mat A \x_{w}
\end{align}
where $\mat A$ is a parameterized diagonal matrix. The weight is high if the word is strongly related to at least one candidate entity. We often observe that uninformative words (e.g. similar to stop words) receive non-negligible scores which add undesired noise to our local context model. As a consequence, we (hard) prune to the top $R \le K$ words with the highest scores\footnote{We implement this in a differentiable way by setting the lowest K-R attention weights in $\bf u$ to $-\infty$ and applying a vanila softmax on top of them. We used the layers Threshold and TemporalDynamicKMaxPooling from Torch nn package, which allow subgradient computation.} and apply a soft-max function on these weights. Define the reduced context:
\begin{align}
\bar c = \{w \in c | u(w) \in \text{topR}(\bf u) \}
\end{align}

Then, the final attention weights are explicitly 
\begin{align}
\beta(w) = 
\begin{cases}
\frac{
	  \exp[u(w)]
  }{
  	\sum_{v \in \bar c} \exp[u(v)]
  } \,. & \text{if $w \in\bar c$}\\
0 & \text{otherwise.}
\end{cases}
\end{align} 
Finally, we define a $\beta$-weighted context-based entity-mention score via
\begin{align}
\Psi(e,c) =  \sum_{w \in \bar c}  \beta(w) \; \x_{e}^\top \mat B \, \x_{w}
\label{eq:psi}
\end{align}
where $\mat B$ is another trainable diagonal matrix. We will later use the same architecture for the \textit{unary} scores of our global ED model.

\noindent{\bf Local Score Combination.}

We integrate these context scores with the context-independent scores encoded in $\hat{p}(e|m)$. Our final (unnormalized) local model is a combination of both $\Psi(e,c)$ and $\log \hat{p}(e|m)$:
\begin{align}
\Psi(e, m, c) = f(\Psi(e, c), \log \hat{p}(e|m))
\label{eq:f}
\end{align}
We find a flexible choice for $f$ to be important and superior to a na\"ive weighted average combination model. We therefore use a neural network with two fully connected layers of 100 hidden units and ReLU non-linearities, which we regularize as suggested in~\cite{denton2015user} by constraining the sum of squares of all weights in the linear layer. We use standard projected SGD for training. The same network is also used in Section~\ref{global}.

Prediction is done independently for each mention $m_i$ and context $c_i$ by maximizing the $\Psi(e, m_i, c_i)$ score.

\noindent{\bf Learning the Local Model.} 

Entity and word embeddings are pre-trained as discussed in Section~\ref{ent_vecs}. Thus, the only learnable parameters are the diagonal matrices $\mat A$ and $\mat B$, plus the parameters of $f$. Having few parameters helps to avoid overfitting and to be able to train with little annotated data. We assume that a set of known mention-entity pairs $\{(m,e^*)\}$ with their respective context windows have been extracted from a corpus. For model fitting, we then utilize a max-margin loss that ranks ground truth entities higher than other candidate entities. This leads us to the objective:
\begin{align}
& \theta^* = \argmin_{\theta} \sum_{D \in \mathcal{D}} \sum_{m \in D} \sum_{e \in \Gamma(m)} g(e,m),  \\
& g(e,m) := [\gamma - \Psi(e^*, m, c) + \Psi(e,m,c)]_+
\nonumber 
\end{align}
where $\gamma>0$ is a margin parameter and $\mathcal{D}$ is a training set of entity annotated documents. We aim to find a $\Psi$ (i.e.~parameterized by $\theta$) such that the score of the correct entity $e^*$ referenced by $m$  is at least a margin $\gamma$ higher than that of any other candidate entity $e$. Whenever this is not the case, the margin violation becomes the experienced loss. 


\section{Document-Level Deep Model}\label{global}

\begin{figure*}
\includegraphics[width=1\textwidth]{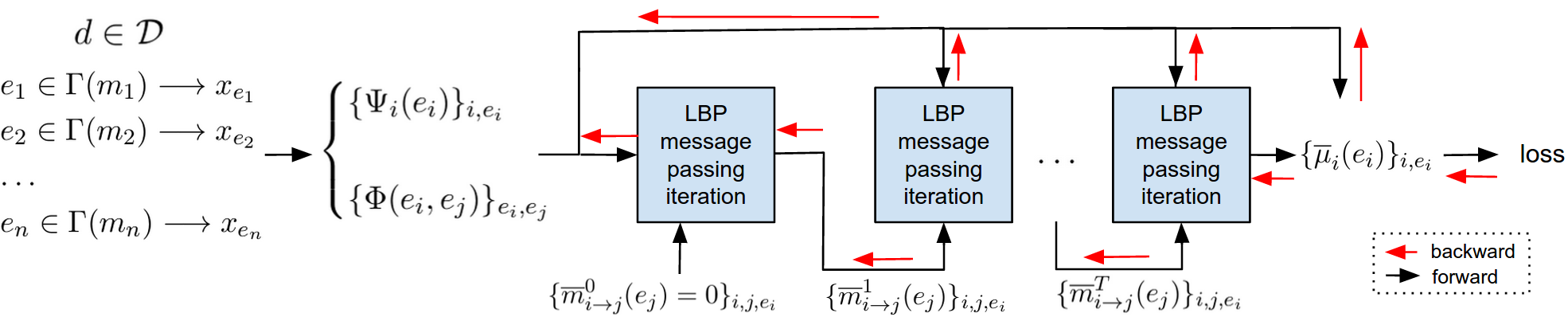}
\caption{Global model: unrolled LBP deep network that is end-to-end differentiable and trainable.}
\label{fig:global-model}
\end{figure*}

Next, we address global ED assuming document coherence among entities. We therefore introduce the notion of a document as consisting of a set of mentions $\m = m_1, \dots, m_n$, along with their context windows $\cc = c_1,\dots c_n$. Our goal is to define a joint probability distribution over $\Gamma(m_1) \times \dots \times \Gamma(m_n) \ni \e$. Each such $\mat e$ selects one candidate entity for each mention in the document. Obviously, the state space of $\e$ grows exponentially in the number of mentions $n$. 

\noindent{\bf CRF Model.} 

Our model is a fully-connected pairwise conditional random field, defined on the log scale as
\begin{align}
\!g(\e,\m,\cc) \!=\! \! \sum_{i=1}^n \Psi_i(e_i) + \! \sum_{i < j} \Phi(e_i,e_j) 
\end{align}
The unary factors are the local scores $\Psi_i(e_i) = \Psi(e_i, c_i)$ described in Eq.~\eqref{eq:psi}. The pairwise factors are bilinear forms of the entity embeddings
\begin{align}
\Phi(e,e') = \frac{2}{n-1} \,\, \x_{e}^{\top}\mat C\, \x_{e'}\,,
\end{align}
where $\mat C$ is a diagonal matrix. Similar to~\cite{ganea2016probabilistic}, the above normalization helps balancing the unary and pairwise terms across documents with different numbers of mentions. 

The function value $g(\e,\m,\cc)$ is supposedly high for semantically related sets of entities that also have local support. The goal of a global ED prediction method is to perform maximum-a-posteriori on this CRF to find the set of entities $\e$ that maximize $g(\e,\m,\cc)$.

\noindent{\bf Differentiable Inference.} 

Training and prediction in binary CRF models as the one above is NP-hard. Therefore, in learning one usually maximizes a likelihood approximation and during operations (i.e.~in prediction) one may use an approximate inference procedure, often based on message-passing. Among many challenges of these approaches, it is worth pointing out that weaknesses of the approximate inference procedure are generally not captured during learning. Inspired by~\cite{domke2011parameter,domke2013learning}, we use~\textit{truncated fitting} of loopy belief propagation (LBP) to a fixed number of message passing iterations. Our model directly optimizes the marginal likelihoods, using the same networks for learning and prediction. As noted by~\cite{domke2013learning}, this method is robust to model mis-specification, avoids inherent difficulties of partition functions and is faster compared to double-loop likelihood training (where, for each stochastic update, inference is run until convergence is achieved). 

Our architecture is shown in Figure~\ref{fig:global-model}. A neural network with $T$ layers encodes $T$ message passing iterations of synchronous max-product LBP\footnote{Sum-product and mean-field performed worse in our experiments.} which is designed to find the most likely (MAP) entity assignments that maximize $g(\e,\m,\cc)$. We also use message damping, which is known to speed-up and stabilize convergence of message passing. Formally, in iteration $t$, mention $m_i$ votes for entity candidate $e \in \Gamma(m_j)$ of mention $m_j$ using the normalized log-message $\overline{m}_{i \rightarrow j}^t(e)$ computed as:
\begin{align}
m_{i \rightarrow j}^{t+1}(e)  = \max_{e' \in \Gamma(m_i)} & \left\{ \Psi_i(e') + \Phi(e,e') \right. \nonumber \\
&  + \sum_{k \neq j} \overline{m}_{k \rightarrow i}^t(e') \}\,.
\label{eq:unnorm-msg}
\end{align}
Herein the first part just reflects the CRF potentials, whereas the second part is defined as 
\begin{align}
\label{eq:norm-msg}
\overline{m}_{i \rightarrow j}^t(e) =\log &[ \delta \cdot \mathrm{softmax}(m_{i \rightarrow j}^{t}(e))\\
                                     & +(1 - \delta) \cdot \exp(\overline{m}_{i \rightarrow j}^{t-1}(e))] \nonumber
\end{align}
where $\delta \in (0,1]$ is a damping factor. Note that, without loss of generality, we simplify the LBP procedure by dropping the factor nodes. The messages at first iteration (layer) are set to zero.

After $T$ iterations (network layers), the beliefs (marginals) are computed as:
\begin{align}
\mu_i(e) & = \Psi_i(e) + \sum_{k \neq i} \overline{m}_{k \rightarrow i}^T(e)\\
\overline{\mu}_i(e) & = \frac{\exp[\mu_i(e)]}{\sum_{e' \in \Gamma(m_i)} \exp[\mu_i(e')]}
\label{eq:marginal}
\end{align}

\begin{figure}[t]
\centering
\includegraphics[width=0.33\textwidth]{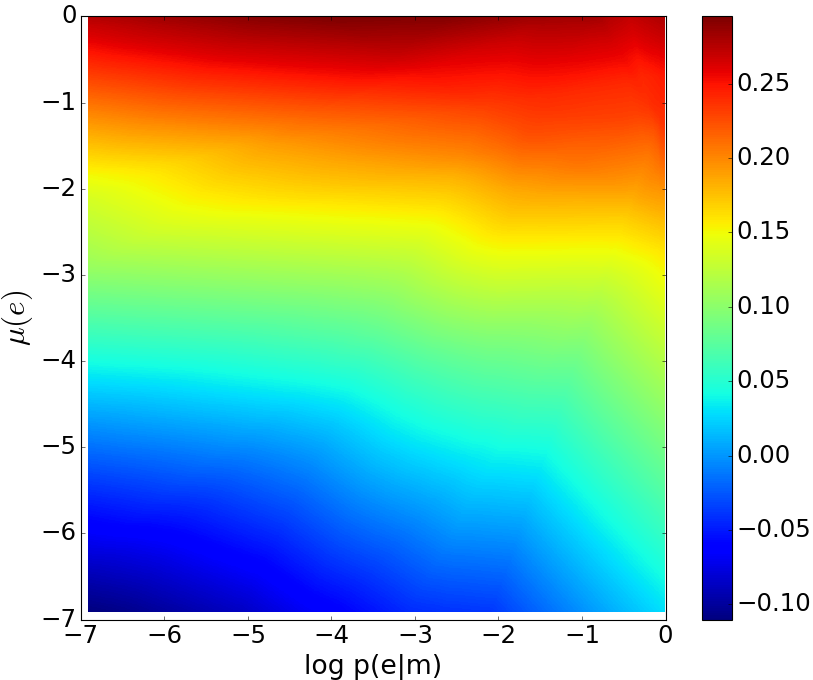}
\caption{Non-linear scoring function of the belief and mention prior learned with a neural network. Achieves a 1.7\% improvement on AIDA-B dataset compared to a weighted average scheme.}
\label{fig:heatmap}
\end{figure}

Similar to the local case, we obtain accuracy improvement when combining the mention-entity prior $\hat{p}(e|m)$ with marginal $\mu_i(e)$ using the same non-linear combination function $f$ from Equation~\ref{eq:f} as follows:
\begin{align}
\rho_i(e) &:= f(\overline{\mu}_i(e), \log \hat{p}(e|m_i))
\end{align}
The learned function $f$ for global ED is non-trivial (see Figure~\ref{fig:heatmap}), showing that the influence of the prior tends to weaken for larger $\mu(e)$, whereas it has a dominating influence whenever the document-level evidence is weak. We also experimented with the prior integrated directly inside the unary factors $\Psi_i(e_i)$, but results were worse because, in some cases, the global entity interaction is not able to recover from strong incorrect priors (e.g. country names have a strong prior towards the respective countries as opposed to national sports teams).

Parameters of our global model are the diagonal matrices $\mat A, \mat B, \mat C$ and the weights of the $f$ network. As before, we find a margin based objective to be the most effective and we suggest to fit parameters by minimizing a ranking loss\footnote{Optimizing a marginal log-likelihood loss function performed worse.} defined as:
\begin{align}
L(\theta) & = \sum_{D \in \mathcal{D}} \sum_{m_i \in D} \sum_{e\in \Gamma(m_i)} h(m_i,e) \\
h(m_i,e) &  = \left[ \gamma - \rho_i(e_i^*) + \rho_i(e) \right]_+ 
\end{align}
Computing this objective is trivial by running $T$ times the steps described by Eqs.~\eqref{eq:unnorm-msg}, \eqref{eq:norm-msg}, followed in the end by the step in Eq.~\eqref{eq:marginal}. Each step is differentiable and the gradient of the model parameters can be computed on the resulting marginals and back-propagated over messages using chain rule. 

At test time, marginals $\rho_i(e)$ are computed jointly per document using this network, but prediction is done independently for each mention $m_i$ by maximizing its respective marginal score.


\begin{table}
\small
\centering
\setlength{\tabcolsep}{1.7pt}
\begin{tabular}{|c|c|c|c|c|}
\hline 
\backslashbox{Method}{Metric} & \tiny{NDCG@1} & \tiny{NDCG@5} & \tiny{NDCG@10} & \tiny{MAP} \\ 
\hline
WikiLinkMeasure (WLM) & 0.54 & 0.52 & 0.55 & 0.48 \\
\hline
\begin{tabular}{@{}c@{}}\cite{yamada2016joint} \\ d = 500\end{tabular} & 0.59 & 0.56 & 0.59 & 0.52 \\
\hline
\begin{tabular}{@{}c@{}}our (canonical pages) \\  d = 300\end{tabular} & 0.624 & 0.589 & 0.615 & 0.549 \\
\hline
\begin{tabular}{@{}c@{}}our (canonical\&hyperlinks) \\ d = 300\end{tabular} & \bf 0.632 & \bf 0.609 & \bf 0.641 & \bf 0.578 \\
\hline
\end{tabular}
\caption{\label{tab:entity-rltd} Entity relatedness results on the test set of~\cite{ceccarelli2013learning}. WLM is a well-known similarity method of~\cite{milne2008learning}.}
\end{table}

\begin{table}
\centering
\setlength{\tabcolsep}{2.5pt}
\small
\begin{tabular}{ccc}
\begin{tabular}{|c|c|c|c|}
\hline
{\bf Dataset} & {\bf  \begin{tabular}{@{}c@{}}Number \\ mentions\end{tabular}} & {\bf  \begin{tabular}{@{}c@{}}Number \\ docs\end{tabular}} & {\bf  \begin{tabular}{@{}c@{}}Mentions \\ per doc\end{tabular}}\\\hline
AIDA-train & 18448 & 946 & 19.5 \\
AIDA-A (valid) & 4791  & 216 & 22.1\\
AIDA-B (test) & 4485  & 231 & 19.4\\ 
\hline
MSNBC & 656  & 20 & 32.8\\ 
AQUAINT & 727  & 50 & 14.5\\
ACE2004 & 257  & 36 & 7.1\\ 
\hline
WNED-CWEB & 11154  & 320 & 34.8\\ 
WNED-WIKI & 6821  & 320 & 21.3\\\hline
\end{tabular}& 
\begin{tabular}{|c|c|}
\hline
{\bf  \begin{tabular}{@{}c@{}}Gold \\ recall\end{tabular}} \\\hline
- \\
96.9\%\\
98.2\%\\ 
\hline
98.5\%\\ 
94.2\%\\
90.6\%\\ 
\hline
91.1\%\\ 
92\%  \\\hline
\end{tabular}
\end{tabular}
\caption{Statistics of ED datasets. \textit{Gold recall} is the percentage of mentions for which the entity candidate set contains the ground truth entity. We only train on mentions with at least one candidate.}\label{tab:datasets}
\end{table}

\section{Candidate Selection}\label{candidate}
We use a mention-entity prior $\hat{p}(e|m)$ both as a feature and for entity candidate selection. It is computed by averaging probabilities from two indexes build from mention entity hyperlink count statistics from Wikipedia and a large Web corpus~\cite{spitkovsky2012cross}. Moreover, we add the YAGO dictionary of~\cite{hoffart2011robust}, where each candidate receives a uniform prior.

Candidate selection, i.e. construction of $\Gamma(e)$, is done for each input mention as follows: first, the top 30 candidates are selected based on the prior $\hat{p}(e|m)$. Then, in order to optimize for memory and run time (LBP has complexity quadratic in S), we keep only $7$ of these entities based on the following heuristic: (i) the top 4 entities based on  $\hat{p}(e|m)$ are selected, (ii) the top 3 entities based on the local context-entity similarity measured using the function from Eq.~\ref{eq:psi} are selected.\footnote{We have used a simpler context vector here computed by simply averaging all its constituent word vectors.}. We refrain from annotating mentions without any candidate entity, implying that precision and recall can be different in our case.

In a few cases, generic mentions of persons (e.g.~"Peter") are coreferences of more specific mentions (e.g.~
"Peter Such") from the same document. We employ a simple heuristic to address this issue: for each mention $m$, if there exist mentions of persons that contain $m$ as a continuous subsequence of words, then we consider the merged set of the candidate sets of these specific mentions as the candidate set for the mention $m$. We decide that a mention refers to a person if its most probable candidate by $\hat{p}(e|m)$ is a person.

\section{Experiments}\label{sec:exp}


\subsection{ED Datasets}\label{sec:datasets}
We validate our ED models on some of the most popular available datasets used by our predecessors\footnote{TAC-KBP datasets used by~\cite{yamada2016joint,globerson2016collective,sun2015modeling} are no longer available. }. We provide statistics in  Table~\ref{tab:datasets}.
\begin{itemize}
\item AIDA-CoNLL dataset~\cite{hoffart2011robust} is one of the biggest manually annotated ED datasets. It contains  training (AIDA-train), validation (AIDA-A) and test (AIDA-B) sets. 
\item MSNBC (MSB), AQUAINT (AQ) and ACE2004 (ACE) datasets cleaned and updated by~\cite{guorobust}\footnote{Available at: \url{bit.ly/2gnSBLg}}
\item WNED-WIKI (WW) and WNED-CWEB (CWEB): are larger, but automatically extracted, thus less reliable. Are built from the ClueWeb and Wikipedia corpora by~\cite{guorobust,gabrilovich2013facc1}.
\end{itemize}


\begin{table}[t]
\small
\centering
\setlength{\tabcolsep}{1.7pt}
\begin{tabular}{|c|c|}
\hline 
Methods & AIDA-B \\ 
\hline
\textit{Local models} & \\
prior $\hat{p}(e|m)$ & 71.9 \\ 
\cite{lazic2015plato} & 86.4 \\
\cite{globerson2016collective} & 87.9 \\
\cite{yamada2016joint} & 87.2 \\
our (local, K=100, R=50) & \bf 88.8 \\
\hline
\hline
\textit{Global models} & \\
\cite{huang2015leveraging} & 86.6 \\
\cite{ganea2016probabilistic} & 87.6\\
\cite{chisholm2015entity} & 88.7\\
\cite{guorobust} & 89.0 \\
\cite{globerson2016collective} & 91.0 \\
\cite{yamada2016joint} & 91.5 \\
\hline
our (global) & \bf 92.22 $\pm$ 0.14 \\
\hline
\end{tabular}
\caption{\label{tab:aida-results} In-KB accuracy for AIDA-B test set. All baselines use KB+YAGO mention-entity index. For our method we show 95\% confidence intervals obtained over 5 runs.}
\end{table}

\begin{table}
\small
\hspace{-0.3cm}
\centering
\setlength{\tabcolsep}{1.7pt}
\begin{tabular}{|c|c|c|c|c|c|}
\hline 
Global methods & MSB & AQ & ACE & CWEB & WW \\ 
\hline
prior $\hat{p}(e|m)$          & 89.3 & 83.2 & 84.4 & 69.8 & 64.2 \\
\hline
\cite{fang2016entity}         & 81.2 & 88.8 & 85.3  & - & - \\
\cite{ganea2016probabilistic} & 91 & 89.2 & \bf 88.7 & - & - \\
\citep{milne2008learning}      & 78 & 85 & 81 & 64.1 & 81.7 \\
\cite{hoffart2011robust}      & 79 & 56 & 80 & 58.6 & 63 \\
\cite{ratinov2011local}       & 75 & 83 & 82 & 56.2 & 67.2 \\
\cite{cheng2013relational}    & 90 & \bf 90 & 86 & 67.5 & 73.4 \\
\cite{guorobust}              & 92 & 87 & 88 & 77 & \bf 84.5 \\
\hline
our (global)           & \bf \begin{tabular}{@{}c@{}}93.7 \\ $\pm$ 0.1 \end{tabular} & \begin{tabular}{@{}c@{}}88.5 \\ $\pm$ 0.4 \end{tabular} & \bf \begin{tabular}{@{}c@{}}88.5 \\ $\pm$ 0.3 \end{tabular} & \bf \begin{tabular}{@{}c@{}}77.9 \\ $\pm$ 0.1 \end{tabular} & \begin{tabular}{@{}c@{}}77.5 \\ $\pm$ 0.1 \end{tabular} \\
\hline
\end{tabular}
\caption{\label{tab:rest-results} Micro F1 results for other datasets. }
\end{table}

\subsection{Training Details and (Hyper)Parameters}\label{sec:training-details}
We explain training details of our approach. All models are implemented in the Torch framework.

\noindent{\bf Entity Vectors Training \& Relatedness Evaluation.}
For entity embeddings only, we use Wikipedia (Feb 2014) corpus for training. Entity vectors are initialized randomly from a 0-mean normal distribution with standard deviation 1. We first train each entity vector on the entity's Wikipedia canonical description page (title words included) for 400 iterations. Subsequently, Wikipedia hyperlinks of the respective entities are used for learning until validation score (described below) stops improving. In each iteration, 20 positive words, each with 5 negative words, are sampled and used for optimization as explained in Section~\ref{ent_vecs}. We use Adagrad~\cite{duchi2011adaptive} with a learning rate of 0.3. We choose embedding size $d = 300$, pre-trained (fixed) Word2Vec word vectors\footnote{By Word2Vec authors: \url{http://bit.ly/1R9Wsqr}}, $\alpha = 0.6$, $\gamma = 0.1$ and window size of 20 for the hyperlinks. We remove stop words before training. Since our method allows to train the embedding of each entity independently of other entities, we decide for efficiency reasons (and without loss of generality) to learn only the vectors of all entities appearing as mention candidates in all the test datasets described in Sec.~\ref{sec:datasets}, a total of 270000 entities. Training of those takes 20 hours on a single TitanX GPU with 12GB of memory.

We test and validate our entity embeddings on the entity relatedness dataset of~\cite{ceccarelli2013learning}. It contains 3319 and 3673 queries for the test and validation sets. Each query consist of one target entity and up to 100 candidate entities with gold standard binary labels indicating if the two entities are related. The associated task requires ranking of related candidate entities higher than the others. Following previous work, we use different evaluation metrics: normalized discounted cumulative gain (NDCG) and mean average precision (MAP). The validation score used during learning is then the sum of the four metrics showed in Table~\ref{tab:entity-rltd}. We perform candidate ranking based on cosine similarity of entity pairs. 

\begin{table}
\hspace{-0.5cm}
\setlength{\tabcolsep}{1.8pt}
\small
\begin{tabular}{cc}
\includegraphics[width=.5\linewidth]{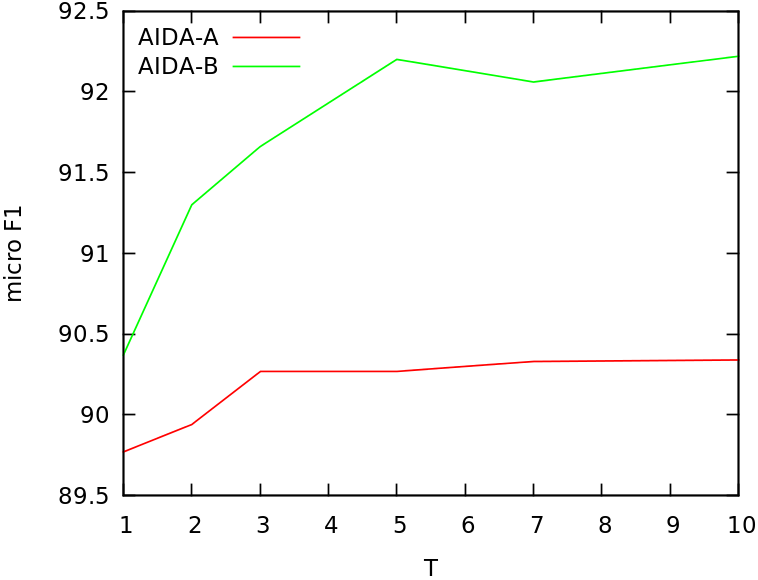} &
\includegraphics[width=.5\linewidth]{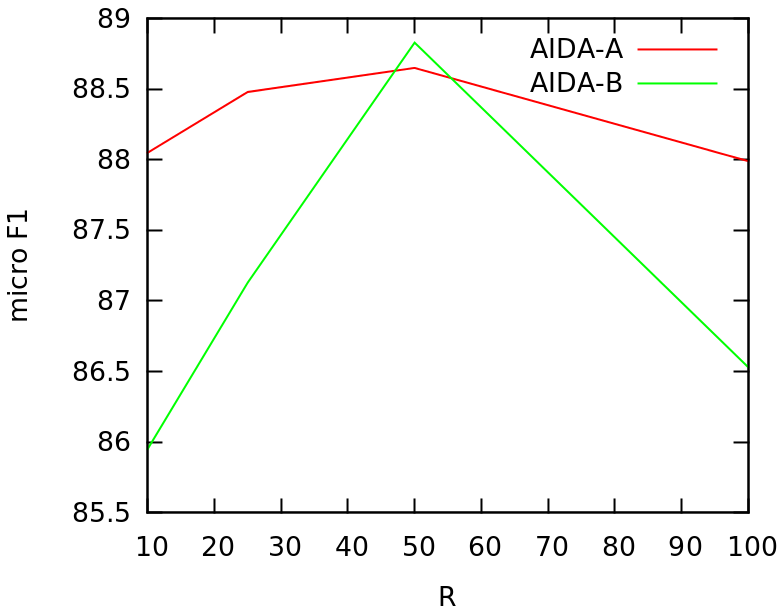}
\end{tabular}
\caption{Effects of two of the hyper-parameters. Left: A low T (e.g.5) is already sufficient for accurate approximate marginals. Right: Hard attention improves accuracy of a local model with K=100.}
\label{tab:plots}
\end{table}

\noindent{\bf Local and Global Model Training.}
Our local and global ED models are trained on AIDA-train (multiple epochs), validated on AIDA-A and tested on AIDA-B and other datasets mentioned in Section~\ref{sec:datasets}. We use Adam~\cite{kingma2014adam} with learning rate of 1e-4 until validation accuracy exceeds 90\%, afterwards setting it to 1e-5. Variable size mini-batches consisting of all mentions in a document are used during training. We remove stop words. Hyper-parameters of the best validated global model are: $\gamma = 0.01, K = 100, R = 25, S = 7, \delta = 0.5, T = 10$. For the local model, $R=50$ was best. Validation accuracy is computed after each 5 epochs. To regularize, we use early stopping, i.e. we stop learning if the validation accuracy does not increase after 500 epochs. Training on a single GPU takes, on average, 2ms per mention, or 16 hours for 1250 epochs over AIDA-train. 

By using diagonal matrices $\mat A, \mat B, \mat C$, we keep the number of parameters very low (approx. 1.2K parameters). This is necessary to avoid overfitting when learning from a very small training set. We also experimented with diagonal plus low-rank matrices, but encountered quality degradation. 


\subsection{Entity Similarity Results}
Results for the entity similarity task are shown in Table~\ref{tab:entity-rltd}. Our method outperforms the well established Wikipedia link measure and the method of~\cite{yamada2016joint} using less information (only word - entity statistics). We note that the best result on this dataset was reported in the unpublished work of~\cite{huang2015leveraging}. Their entity embeddings are trained on many more sources of information (e.g. KG links, relations, entity types). However, our focus was to prove that lightweight trained embeddings useful for the ED task can also perform decently for the entity similarity task. We emphasize that our global ED model outperforms Huang's ED model (Table~\ref{tab:aida-results}), likely due to the power of our local and joint neural network architectures. For example, our attention mechanism clearly benefits from explicitly embedding words and entities in the same space. 


\begin{table}
\setlength{\tabcolsep}{1.8pt}
\small
\begin{tabular}{cc}
\begin{tabular}{|c|c|c|}
\hline
{\bf  \begin{tabular}{@{}c@{}}Freq \\ gold \\ entity\end{tabular}} &  \begin{tabular}{@{}c@{}}Number \\ mentions\end{tabular} & \begin{tabular}{@{}c@{}}Solved  \\ correctly\end{tabular}\\\hline
0 & 5 & 80.0 \%  \\
1-10 & 0 & - \\
11-20 & 4 & 100.0\%  \\
21-50 & 50 & 90.0\%  \\
$> 50$ & 4345 & 94.2\%  \\
\hline
\end{tabular}& 
\begin{tabular}{|c|c|c|}
\hline
{\bf  \begin{tabular}{@{}c@{}}$\hat{p}(e|m)$ \\ gold \\ entity\end{tabular}} &  \begin{tabular}{@{}c@{}}Number \\ mentions\end{tabular} & \begin{tabular}{@{}c@{}}Solved  \\ correctly\end{tabular} \\\hline
$\leq 0.01$ & 36 & 89.19\%  \\
0.01 - 0.03 & 249 & 88.76\%  \\
0.03 - 0.1 & 306 & 82.03\%  \\
0.1 - 0.3 & 381 & 86.61\%  \\
$> 0.3$ & 3431 & 96.53\%  \\
\hline
\end{tabular}
\end{tabular}
\caption{ED accuracy on AIDA-B for our best system splitted by Wikipedia hyperlink frequency and mention prior of the ground truth entity, in cases where the gold entity appears in the candidate set. }
\label{tab:freq}
\end{table}

\subsection{ED Baselines \& Results}
We compare with systems that report state-of-the-art results on the datasets. Some baseline scores from Table~\ref{tab:rest-results} are taken from~\cite{guorobust}. The best results for the AIDA datasets are reported by~\cite{yamada2016joint} and~\cite{globerson2016collective}. We do not compare against~\cite{pershina2015personalized} since, as noted also by~\cite{globerson2016collective}, their mention  index artificially includes the gold entity (guaranteed gold recall), which is not a realistic setting. 

For a fair comparison with prior work, we use in-KB accuracy and micro F1 (averaged per mention) metrics to evaluate our approach. Results are shown in Tables~\ref{tab:aida-results} and~\ref{tab:rest-results}. We run our system 5 times, each time we pick the best model on the validation set, and report results on the test set for these models. We obtain state of the art accuracy on AIDA which is the largest and hardest (by the accuracy of the $\hat{p}(e|m)$ baseline) manually created ED dataset . We are also competitive on the other datasets. It should be noted that all the other methods use, at least partially, engineered features. The merit of our proposed method is to show that, with the exception of the $\hat{p}(e|m)$ feature, a neural network is able to learn the best features for ED without requiring expert input.

To gain further insight, we analyzed the accuracy on the AIDA-B dataset for situations where gold entities have low frequency or mention prior. Table~\ref{tab:freq} shows that our method performs well in these harder cases.


\begin{table*}[t]
\small
\centering
\setlength{\tabcolsep}{2.7pt}
\begin{tabular}{|c|c|c|C{9.5cm}|}
\hline 
\bf Mention & Gold entity & \bf \begin{tabular}{@{}c@{}}$\hat{p}(e|m)$ \\ of gold \\ entity\end{tabular} & \bf Attended contextual words\\ 
\hline
{\color{green}Scotland} & \begin{tabular}{@{}c@{}}{\color{blue}Scotland national}  \\ {\color{blue}rugby union team}  \end{tabular}  & 0.034 & England Rugby team squad Murrayfield Twickenham national play Cup Saturday World game George following Italy week Friday selection dropped row month \\
\hline
{\color{green}Wolverhampton } & \begin{tabular}{@{}c@{}}{\color{blue}Wolverhampton}  \\ {\color{blue}Wanderers F.C.}  \end{tabular}   & 0.103 &  matches League Oxford Hull league Charlton Oldham Cambridge Sunderland Blackburn Sheffield Southampton Huddersfield Leeds Middlesbrough Reading Coventry Darlington Bradford Birmingham Enfield  Barnsley\\
\hline
{\color{green} Montreal } & {\color{blue} Montreal Canadiens} & 0.021 &  League team Hockey Toronto Ottawa games Anaheim Edmonton Rangers Philadelphia Caps Buffalo Pittsburgh Chicago Louis National home Friday York Dallas Washington Ice\\
\hline
{\color{green} Santander} & {\color{blue} Santander Group} & 0.192 & Carlos Telmex Mexico Mexican group firm market week Ponce debt shares buying Televisa earlier pesos share stepped Friday analysts ended 
 \\
\hline
{\color{green} World Cup} & \begin{tabular}{@{}c@{}}{\color{blue}FIS Alpine}  \\ {\color{blue}Ski World Cup}  \end{tabular} & 0.063 & Alpine ski national slalom World  Skiing Whistler downhill Cup events race consecutive weekend Mountain Canadian racing 
 \\
\hline
\end{tabular}
\caption{Examples of context words selected by our local attention mechanism. Distinct words are sorted decreasingly by attention weights and only words with non-zero weights are shown. }
\label{tab:local-words} 
\end{table*}


\subsection{Hyperparameter Studies}\label{sec:hyp-studies}

In Table~\ref{tab:plots}, we analyze the effect of two hyper-parameters. First, we see that hard attention (i.e. $R < K$) helps reducing the noise from uninformative context words (as opposed to keeping all words when $R = K$). 

Second, we see that a small number of LBP iterations (hard-coded in our network) is enough to obtain good accuracy. This speeds up training and testing compared to traditional methods that run LBP until convergence. An explanation is that a truncated version of LBP can perform well enough if used at both training and test time.


\subsection{Qualitative Analysis of Local Model}

In Table~\ref{tab:local-words} we show some examples of context words attended by our local model for correctly solved hard cases (where the mention prior of the correct entity is low). One can notice that words relevant for at least one entity candidate are chosen by our model in most of the cases.


\subsection{Error Analysis}

We analyse some of the errors made by our model on the AIDA-B dataset. We mostly observe three situations: i) annotation errors, ii) gold entities that do not appear in mentions' candidate sets, or iii) gold entities with very low $p(e|m)$ prior whose mentions have an incorrect entity candidate with high prior. For example, the mention "Italians" refers in some specific context to the entity "Italy national football team" rather than the entity representing the country. The contextual information is not strong enough in this case to avoid an incorrect prediction. On the other hand, there are situations where the context can be misleading, e.g. a document heavily discussing about cricket will favor resolving the mention "Australia" to the entity "Australia national cricket team" instead of the gold entity "Australia" (naming a location of cricket games in the given context). 


\section{Conclusion}
We have proposed a novel deep learning architecture for entity disambiguation that combines entity embeddings, a contextual attention mechanism, an adaptive local score combination, as well as unrolled differentiable message passing for global inference.  Compared to many other methods, we do not rely on hand-engineered features, nor on an extensive corpus for entity co-occurrences or relatedness.  Our system is fully differentiable, although we chose to pre-train word and entity embeddings.  Extensive experiments show the competitiveness of our approach across a wide range of corpora. In the future, we would like to extend this system to perform nil detection, coreference resolution and mention detection.

Our code and data are publicly available: \url{http://github.com/dalab/deep-ed}

\section*{Acknowledgments}

We thank Aurelien Lucchi, Marina Ganea, Jason Lee, Florian Schmidt and Hadi Daneshmand for their comments and suggestions.

This research was supported by the Swiss National Science Foundation (SNSF) grant number 407540$\_$167176 under the project "Conversational Agent for Interactive Access to Information".


\bibliography{emnlp2017}

\begin{thebibliography}{32}
\expandafter\ifx\csname natexlab\endcsname\relax\def\natexlab#1{#1}\fi

\bibitem[{Ceccarelli et~al.(2013)Ceccarelli, Lucchese, Orlando, Perego, and
  Trani}]{ceccarelli2013learning}
Diego Ceccarelli, Claudio Lucchese, Salvatore Orlando, Raffaele Perego, and
  Salvatore Trani. 2013.
\newblock Learning relatedness measures for entity linking.
\newblock In \emph{Proceedings of the 22nd ACM international conference on
  Information \& Knowledge Management}, pages 139--148. ACM.

\bibitem[{Cheng and Roth(2013)}]{cheng2013relational}
Xiao Cheng and Dan Roth. 2013.
\newblock Relational inference for wikification.
\newblock \emph{Urbana}, 51(61801):16--58.

\bibitem[{Chisholm and Hachey(2015)}]{chisholm2015entity}
Andrew Chisholm and Ben Hachey. 2015.
\newblock Entity disambiguation with web links.
\newblock \emph{Transactions of the Association for Computational Linguistics},
  3:145--156.

\bibitem[{Denton et~al.(2015)Denton, Weston, Paluri, Bourdev, and
  Fergus}]{denton2015user}
Emily Denton, Jason Weston, Manohar Paluri, Lubomir Bourdev, and Rob Fergus.
  2015.
\newblock User conditional hashtag prediction for images.
\newblock In \emph{Proceedings of the 21th ACM SIGKDD International Conference
  on Knowledge Discovery and Data Mining}, pages 1731--1740. ACM.

\bibitem[{Domke(2011)}]{domke2011parameter}
Justin Domke. 2011.
\newblock Parameter learning with truncated message-passing.
\newblock In \emph{Computer Vision and Pattern Recognition (CVPR), 2011 IEEE
  Conference on}, pages 2937--2943. IEEE.

\bibitem[{Domke(2013)}]{domke2013learning}
Justin Domke. 2013.
\newblock Learning graphical model parameters with approximate marginal
  inference.
\newblock \emph{IEEE transactions on pattern analysis and machine
  intelligence}, 35(10):2454--2467.

\bibitem[{Duchi et~al.(2011)Duchi, Hazan, and Singer}]{duchi2011adaptive}
John Duchi, Elad Hazan, and Yoram Singer. 2011.
\newblock Adaptive subgradient methods for online learning and stochastic
  optimization.
\newblock \emph{Journal of Machine Learning Research}, 12(Jul):2121--2159.

\bibitem[{Fang et~al.(2016)Fang, Zhang, Wang, Chen, and Li}]{fang2016entity}
Wei Fang, Jianwen Zhang, Dilin Wang, Zheng Chen, and Ming Li. 2016.
\newblock Entity disambiguation by knowledge and text jointly embedding.
\newblock \emph{CoNLL 2016}, page 260.

\bibitem[{Ferragina and Scaiella(2010)}]{ferragina2010tagme}
Paolo Ferragina and Ugo Scaiella. 2010.
\newblock Tagme: on-the-fly annotation of short text fragments (by wikipedia
  entities).
\newblock In \emph{Proceedings of the 19th ACM international conference on
  Information and knowledge management}, pages 1625--1628. ACM.

\bibitem[{Francis-Landau et~al.(2016)Francis-Landau, Durrett, and
  Klein}]{francis2016capturing}
Matthew Francis-Landau, Greg Durrett, and Dan Klein. 2016.
\newblock Capturing semantic similarity for entity linking with convolutional
  neural networks.
\newblock \emph{arXiv preprint arXiv:1604.00734}.

\bibitem[{Gabrilovich et~al.(2013)Gabrilovich, Ringgaard, and
  Subramanya}]{gabrilovich2013facc1}
Evgeniy Gabrilovich, Michael Ringgaard, and Amarnag Subramanya. 2013.
\newblock Facc1: Freebase annotation of clueweb corpora, version 1 (release
  date 2013-06-26, format version 1, correction level 0).
\newblock \emph{Note: http://lemurproject. org/clueweb09/FACC1/Cited by}, 5.

\bibitem[{Ganea et~al.(2016)Ganea, Ganea, Lucchi, Eickhoff, and
  Hofmann}]{ganea2016probabilistic}
Octavian-Eugen Ganea, Marina Ganea, Aurelien Lucchi, Carsten Eickhoff, and
  Thomas Hofmann. 2016.
\newblock Probabilistic bag-of-hyperlinks model for entity linking.
\newblock In \emph{Proceedings of the 25th International Conference on World
  Wide Web}, pages 927--938. International World Wide Web Conferences Steering
  Committee.

\bibitem[{Globerson et~al.(2016)Globerson, Lazic, Chakrabarti, Subramanya,
  Ringgaard, and Pereira}]{globerson2016collective}
Amir Globerson, Nevena Lazic, Soumen Chakrabarti, Amarnag Subramanya, Michael
  Ringgaard, and Fernando Pereira. 2016.
\newblock Collective entity resolution with multi-focal attention.
\newblock In \emph{ACL (1)}.

\bibitem[{Guo and Barbosa(2016)}]{guorobust}
Zhaochen Guo and Denilson Barbosa. 2016.
\newblock Robust named entity disambiguation with random walks.

\bibitem[{He et~al.(2013)He, Liu, Li, Zhou, Zhang, and Wang}]{he2013learning}
Zhengyan He, Shujie Liu, Mu~Li, Ming Zhou, Longkai Zhang, and Houfeng Wang.
  2013.
\newblock Learning entity representation for entity disambiguation.
\newblock In \emph{ACL (2)}, pages 30--34.

\bibitem[{Hoffart et~al.(2011)Hoffart, Yosef, Bordino, F{\"u}rstenau, Pinkal,
  Spaniol, Taneva, Thater, and Weikum}]{hoffart2011robust}
Johannes Hoffart, Mohamed~Amir Yosef, Ilaria Bordino, Hagen F{\"u}rstenau,
  Manfred Pinkal, Marc Spaniol, Bilyana Taneva, Stefan Thater, and Gerhard
  Weikum. 2011.
\newblock Robust disambiguation of named entities in text.
\newblock In \emph{Proceedings of the Conference on Empirical Methods in
  Natural Language Processing}, pages 782--792. Association for Computational
  Linguistics.

\bibitem[{Huang et~al.(2015)Huang, Heck, and Ji}]{huang2015leveraging}
Hongzhao Huang, Larry Heck, and Heng Ji. 2015.
\newblock Leveraging deep neural networks and knowledge graphs for entity
  disambiguation.
\newblock \emph{arXiv preprint arXiv:1504.07678}.

\bibitem[{Ji(2016)}]{ji2016}
Heng Ji. 2016.
\newblock \href {http://nlp.cs.rpi.edu/kbp/2014/elreading.html} {Entity
  discovery and linking reading list}.

\bibitem[{Kingma and Ba(2014)}]{kingma2014adam}
Diederik Kingma and Jimmy Ba. 2014.
\newblock Adam: A method for stochastic optimization.
\newblock \emph{arXiv preprint arXiv:1412.6980}.

\bibitem[{Lafferty et~al.(2001)Lafferty, McCallum, Pereira
  et~al.}]{lafferty2001conditional}
John Lafferty, Andrew McCallum, Fernando Pereira, et~al. 2001.
\newblock Conditional random fields: Probabilistic models for segmenting and
  labeling sequence data.
\newblock In \emph{Proceedings of the eighteenth international conference on
  machine learning, ICML}, volume~1, pages 282--289.

\bibitem[{Lazic et~al.(2015)Lazic, Subramanya, Ringgaard, and
  Pereira}]{lazic2015plato}
Nevena Lazic, Amarnag Subramanya, Michael Ringgaard, and Fernando Pereira.
  2015.
\newblock Plato: A selective context model for entity resolution.
\newblock \emph{Transactions of the Association for Computational Linguistics},
  3:503--515.

\bibitem[{Mikolov et~al.(2013)Mikolov, Sutskever, Chen, Corrado, and
  Dean}]{mikolov2013distributed}
Tomas Mikolov, Ilya Sutskever, Kai Chen, Greg~S Corrado, and Jeff Dean. 2013.
\newblock Distributed representations of words and phrases and their
  compositionality.
\newblock In \emph{Advances in neural information processing systems}, pages
  3111--3119.

\bibitem[{Milne and Witten(2008)}]{milne2008learning}
David Milne and Ian~H Witten. 2008.
\newblock Learning to link with wikipedia.
\newblock In \emph{Proceedings of the 17th ACM conference on Information and
  knowledge management}, pages 509--518. ACM.

\bibitem[{Murphy et~al.(1999)Murphy, Weiss, and Jordan}]{murphy1999loopy}
Kevin~P Murphy, Yair Weiss, and Michael~I Jordan. 1999.
\newblock Loopy belief propagation for approximate inference: An empirical
  study.
\newblock In \emph{Proceedings of the Fifteenth conference on Uncertainty in
  artificial intelligence}, pages 467--475. Morgan Kaufmann Publishers Inc.

\bibitem[{Pennington et~al.(2014)Pennington, Socher, and
  Manning}]{pennington2014glove}
Jeffrey Pennington, Richard Socher, and Christopher~D Manning. 2014.
\newblock Glove: Global vectors for word representation.
\newblock In \emph{EMNLP}, volume~14, pages 1532--43.

\bibitem[{Pershina et~al.(2015)Pershina, He, and
  Grishman}]{pershina2015personalized}
Maria Pershina, Yifan He, and Ralph Grishman. 2015.
\newblock Personalized page rank for named entity disambiguation.

\bibitem[{Ratinov et~al.(2011)Ratinov, Roth, Downey, and
  Anderson}]{ratinov2011local}
Lev Ratinov, Dan Roth, Doug Downey, and Mike Anderson. 2011.
\newblock Local and global algorithms for disambiguation to wikipedia.
\newblock In \emph{Proceedings of the 49th Annual Meeting of the Association
  for Computational Linguistics: Human Language Technologies-Volume 1}, pages
  1375--1384. Association for Computational Linguistics.

\bibitem[{Spitkovsky and Chang(2012)}]{spitkovsky2012cross}
Valentin~I Spitkovsky and Angel~X Chang. 2012.
\newblock A cross-lingual dictionary for english wikipedia concepts.

\bibitem[{Sukhbaatar et~al.(2015)Sukhbaatar, Weston, Fergus
  et~al.}]{sukhbaatar2015end}
Sainbayar Sukhbaatar, Jason Weston, Rob Fergus, et~al. 2015.
\newblock End-to-end memory networks.
\newblock In \emph{Advances in neural information processing systems}, pages
  2440--2448.

\bibitem[{Sun et~al.(2015)Sun, Lin, Tang, Yang, Ji, and Wang}]{sun2015modeling}
Yaming Sun, Lei Lin, Duyu Tang, Nan Yang, Zhenzhou Ji, and Xiaolong Wang. 2015.
\newblock Modeling mention, context and entity with neural networks for entity
  disambiguation.
\newblock In \emph{IJCAI}, pages 1333--1339.

\bibitem[{Yamada et~al.(2016)Yamada, Shindo, Takeda, and
  Takefuji}]{yamada2016joint}
Ikuya Yamada, Hiroyuki Shindo, Hideaki Takeda, and Yoshiyasu Takefuji. 2016.
\newblock Joint learning of the embedding of words and entities for named
  entity disambiguation.
\newblock \emph{CoNLL 2016}, page 250.

\bibitem[{Zwicklbauer et~al.(2016)Zwicklbauer, Seifert, and
  Granitzer}]{zwicklbauer2016robust}
Stefan Zwicklbauer, Christin Seifert, and Michael Granitzer. 2016.
\newblock Robust and collective entity disambiguation through semantic
  embeddings.
\newblock In \emph{Proceedings of the 39th International ACM SIGIR conference
  on Research and Development in Information Retrieval}, pages 425--434. ACM.

\end{thebibliography}
\bibliographystyle{emnlp_natbib}

\newpage

\begin{table*}
\small
\centering
\setlength{\tabcolsep}{4pt}
\begin{tabular}{|c|C{11cm}|}
\hline 
\bf Entity & \bf Closest words sorted by cosine similarity \\ 
\hline
{\color{blue}Japan national football team} & Japan player Shizuoka Yokohama played Asian USISL Saitama Okada Nakamura Tokyo Pele  matches Japanese Korea players Tanaka soccer Chunnam game Suwon Takuya Kawaguchi Mizuno match Qatar team Eto Eiji football playing Confederations tournament Kagawa Chiba\\
\hline
{\color{blue}Apple} & apple fruit berry grape varieties apples crop pear potato blueberry strawberry growers peach orchards pears Prunus grower Rubus citrus spinosa tomato berries Blueberry peaches grapes almond juice melon bean apricot insect vegetable strawberries olive pomegranate Vaccinium cherries potatoes Strawberry plums cultivar Apples harvest figs cultivars sunflower beet \\
\hline
{\color{blue}Apple Inc.} & Apple software computer Microsoft Adobe hardware company iPod PC product Dell laptop Mac computers Macintosh Flash video desktop iPhone Digital Windows app PCs Intel technology device iTunes Motorola Sony digital Multimedia iPad HP licensing multimedia Nokia apps smartphone laptops Computer previewed products application Jobs devices startup\\
\hline
{\color{blue}Queen (band)} & U2 band singer Avenged Rockers Coldplay concert Lynyrd Kiss Metallica Killers rerecorded song Beatles rock Stones recording Slash Singer touring musician music CD Dirty Moby rockers Sting Blackest songs rocker\\
\hline
{\color{blue}Germany} &  Germany Berlin German Munich Hamburg Austria Cologne Bavaria Hessen country Europe Wernigerode Saxony western Germans Schwaben Switzerland TuS Heilbronn Realschule Westfalen Deutschland Brandenburg eastern Rudolf Glarus Wolfgang Esslingen Kaserne Swabia Schwerin Andreas Poland Helmut Palatinate history Darmstadt Rhein Harald Ludwigsburg Kiel\\
\hline
{\color{blue}Barack Obama} & Obama campaign President presidential endorsed Democrat Clinton nominee Presidential inauguration Senator senator administration speech Barack Democratic appointee Washington Republican vote Tuesday Secretary election Administration elect nomination Bush November president congressman Senate endorsing announcement candidacy\\
\hline
{\color{blue}Leicestershire} & curacy town Yeomanry Buckinghamshire Leicestershire Bedfordshire Lichfield Wiltshire Shropshire almshouses Lancashire Stonyhurst\\
\hline
{\color{blue}Leicestershire County Cricket Club} & Warwickshire batsman England Hampshire Leicestershire Trott Glamorgan Nottinghamshire Northants Lancashire Middlesex Essex Giles fielding Porterfield Test Surrey cricketer centurion Gough Bevan Sussex Gloucestershire bowled Worcestershire Tests Martyn Croft Derbyshire Clarke overs bowler Lancastrian played Northamptonshire Kent Vaughan Fletcher captaining internationals batting Gilchrist Notts batted cricket\\
\hline
\end{tabular}
\caption{Closest words to a given entity. Words with at least 500 frequency in the Wikipedia corpus are shown.}
\label{tbl:entity-words} 
\end{table*}

\end{document}